\documentclass[journal]{IEEEtran}

\usepackage{hyperref}
\usepackage{multicol}

\usepackage{color}

\ifCLASSINFOpdf
   \usepackage[pdftex]{graphicx}
\else
   \usepackage[dvips]{graphicx}
\fi
\usepackage[cmex10]{amsmath}
\usepackage{hyperref}
\hypersetup{
    colorlinks = true
}
\begin{document}
%
\title{Unsupervised Learning for Computational Phenotyping}

%
%
%

\author{Chris Hodapp (\url{chodapp3@gatech.edu})
}

\maketitle

\begin{abstract}
With large volumes of health care data comes the research area of
computational phenotyping, making use of techniques such as machine
learning to describe illnesses and other clinical concepts from the
data itself.  The ``traditional'' approach of using supervised
learning relies on a domain expert, and has two main limitations:
requiring skilled humans to supply correct labels limits its
scalability and accuracy, and relying on existing clinical
descriptions limits the sorts of patterns that can be found. For
instance, it may fail to acknowledge that a disease treated as a
single condition may really have several subtypes with different
phenotypes, as seems to be the case with asthma and heart
disease. Some recent papers cite successes instead using unsupervised
learning.  This shows great potential for finding patterns in
Electronic Health Records that would otherwise be hidden and that can
lead to greater understanding of conditions and treatments. This work
implements a method derived strongly from Lasko \emph{et al.}, but
implements it in Apache Spark and Python and generalizes it to
laboratory time-series data in MIMIC-III.  It is released as an
open-source tool for exploration, analysis, and visualization,
available at: \url{https://github.com/Hodapp87/mimic3_phenotyping}.
\end{abstract}


%

\section{Introduction \& Background}

The field of \emph{computational phenotyping}\cite{Che2015} has
emerged recently as a way of learning more from the increasing volumes
of Electronic Health Records available, and the volume of this data
ties it in naturally with fields like machine learning and data
mining.  The ``traditional'' approach of supervised learning over
classifications has two noted problems:

\begin{itemize}
\item{It requires the time and attention of that domain expert in
  order to provide classification information over which to train a
  model, and this requirement on human attention limits the amount of
  data available (and, to an extent, its accuracy.)}
\item{It tends to limit the patterns that can be found to what
  existing classifications acknowledge.  If a disease treated as a
  single condition really has multiple subtypes with different
  phenotypes, the model will not reflect this - for instance, asthma
  and heart disease\cite{Lasko2013}.}
\end{itemize}

Some recent papers\cite{Marlin,Lasko2013,Johnson2016} cite
successes with approaches instead using unsupervised learning on
time-series data.  In Lasko \emph{et al.}\cite{Lasko2013}, such an
approach applied to serum uric acid measurements was able to
distinguish gout and acute leukemia with no prior classifications
given in training.  Marlin \emph{et al.}\cite{Marlin} examines 13
physiological measures from a pediatric ICU (such as pulse oximetric
saturation, heart rate, and respiratory rate). Che \emph{et
  al.}\cite{Che2015} likewise uses ICU data, but focuses on certain
ICD-9 codes rather than mortality.

This approach still has technical barriers.  Time-series in healthcare
data frequently are noisy, spare, heterogeneous, or irregularly
sampled, and commonly Gaussian processes are employed here in order to
condition the data into a more regular form as a pre-processing step.
In \cite{Lasko2013}, Gaussian process regression produces a model
which generates a continuous, interpolated time-series providing both
predicted mean and variance, then applies a two-layer stacked sparse
autoencoder (compared with a five-layer stacked denoising autoencoder
in \cite{Che2015}, without Gaussian process regression).

The goal undertaken here was to reimplement a combination of some
earlier results (focusing mainly on that of Lasko \emph{et
  al.}\cite{Lasko2013}) using \href{http://spark.apache.org/}{Apache
  Spark} and the MIMIC-III critical care database\cite{Johnson2016a},
and able to run primarily on a ``standard'' Spark setup such as
\href{https://aws.amazon.com/emr/}{Amazon Elastic MapReduce}.
However, presently it relies on Python in order to use Keras and
scikit-learn for feature learning and t-SNE.

The software behind this work is also released as an open source tool
for accomodating exploration, analysis, and visualization using the
techniques described herein. It is available at:
\url{https://github.com/Hodapp87/mimic3_phenotyping}.


\section{Approach \& Implementation}

The main problems that the implementation tries to address within
these parameters are:

\begin{itemize}
\item{Loading MIMIC-III data into a form usable from Spark}
\item{Identifying relevant laboratory tests, admissions, and ICD-9
  codes on which to focus}
\item{Preprocessing the time-series data with Gaussian process
  regression}
\item{Using a two-layer stacked sparse autoencoder to perform feature
  learning}
\item{Visualizing the new feature space and identifying potential
  clusters}
\end{itemize}

This section describes the general approach, and the Experimental
Evaluation section on page \pageref{expeval} gives specific examples
that were tested.

\subsection{Loading \& Selecting Data}

The MIMIC-III database is supplied as a collection of \texttt{.csv.gz}
files (that is, comma-separated values, compressed with gzip).  By way
of \texttt{spack-csv}, Apache Spark 2.x is able to load these files
natively as tabular data, i.e. a \texttt{DataFrame}.  All work
described here used the following tables\cite{Johnson2016a}:

\begin{itemize}
\item \texttt{LABEVENTS}: Timestamped laboratory measurements for patients
\item \texttt{DIAGNOSES\_ICD}: ICD-9 diagnoses for patients (per-admission)
\item \texttt{D\_ICD\_DIAGNOSES}: Information on each ICD-9 diagnosis
\item \texttt{D\_LABEVENTS}: Information on each type of laboratory event
\end{itemize}

The process requires two ICD-9 categories, and one LOINC code for a
lab test.  Admissions are filtered to those which contain a lab
time-series of the given LOINC code and containing at least 3 samples,
containing either ICD-9 category mutually exclusively (that is, at
least one diagnosis of the first ICD-9 category, but none of the
second, or at least one of the second ICD-9 category, and none of the
first).

As an aid to this process, the tool can produce a matrix in which each
row represents an ICD-9 category (starting with the most occurrences
and limited at some number), each column represents a likewise ordered
LOINC code (\url{http://loinc.org/}), and each intersection contains
the number of admissions with an ICD-9 diagnosis of that category and
a laboratory time-series of that LOINC code.  It does not tell whether
a pair of ICD-9 categories, mutually excluding each other, may produce
enough data, but it may still give a meaningful estimate.  The below
shows an example of this matrix, limited to the top 4 LOINC codes
(incidentally, all blood measurements) and top 12 ICD-9 categories for
space reasons:

\begin{tabular}{l|llllllll}
  \hline
  ICD-9 & \multicolumn{4}{c}{LOINC code}\\
  category & 11555-0 & 11556-8 & 11557-6 & 11558-4 \\
  \cline{2-5}
  427 & 12456 & 12454 & 12458 & 12779 \\
  276 & 11392 & 11393 & 11393 & 11962 \\
  428 & 11198 & 11195 & 11196 & 11515 \\
  401 & 11186 & 11187 & 11188 & 11525 \\
  518 & 11386 & 11387 & 11386 & 11545 \\
  250 & 8238 & 8238 & 8240 & 8574 \\
  414 & 9243 & 9242 & 9242 & 9412 \\
  272 & 7733 & 7733 & 7736 & 7919 \\
  285 & 6423 & 6421 & 6422 & 6720 \\
  584 & 6541 & 6541 & 6542 & 6834 \\
  V45 & 4862 & 4860 & 4861 & 5068 \\
  599 & 3815 & 3816 & 3815 & 3983 \\
  \hline
\end{tabular}

All processing at this stage was done via Spark's DataFrame
operations, aside from the final conversion to an RDD containing
individual time-series.

\subsection{Preprocessing}

\subsubsection{Time Warping}

The covariance function that is used in Gaussian process regression
(and explained after this section) contains a time-scale parameter
$\tau$ which embeds assumptions on how closely correlated nearby
samples are in time.  This value is assumed not to change within the
time-series - that is, it is assumed to be
\textit{stationary}\cite{Lasko2013}.  This assumption is often
incorrect, but under the assumption that more rapidly-varying things
(that is, shorter time-scale) are measured more frequently, an
approximation can be applied to try to make the time-series more
stationary - in the form of changing the distance in time between
every pair of adjacent samples in order to shorten longer distances,
but lengthen shorter ones\cite{Lasko2013}.  For original distance $d$,
the warped distance is $d'=d^{1/a}+b$, using $a=3, b=0$ (these
values were taken directly from equation 5 of \cite{Lasko2013} and not
tuned further).

Thomas Lasko also related in an email that this assumption (that
measurement frequency was proportional to how volatile the thing being
measured is) is not true for all medical tests.  He referred to
another paper of his\cite{Lasko2015} for a more robust approach,
however, this is not used here.

\subsubsection{Gaussian Process Regression}

In order to condition the irregular and sparse time-series data from
the prior step, Gaussian process regression was used.  The method used
here is what Lasko \emph{et al.}\cite{Lasko2013} described, which in
more depth is the method described in algorithm 2.1 of Rasmussen \&
Williams\cite{Rasmussen2004}.

In brief, Gaussian process regression (GPR) is a Bayesian
non-parametric, or less parametric, method of supervised learning over
noisy observations\cite{Ebden2015,Lasko2013,Rasmussen2004}.  It is not
completely free-form, but it infers a function constrained only by the
mean function (which here is assumed to be 0 and can be ignored) and
the covariance function $k(t,t')$ of an underlying
infinite-dimensional Gaussian distribution.  It is not exclusive to
time-series data, but $t$ is used here as in this work GPR is done
only on time-series data.

That covariance function $k$ defines how dependent observations are on
each other, and so a common choice is the squared
exponential\cite{Ebden2015}:

$$k(t,t') = \sigma_n^2\exp\bigg[\frac{-(t-t')^2}{2l^2}\bigg]$$

Note that $k$ approaches a maximum of $\sigma_n^2$ as $t$ and $t'$ are
further, and $k$ approaches a minimum of 0 as $t$ and $t'$ are closer.
Intuitively, this makes sense for many ``natural'' functions: we
expect closer $t$ values to have more strongly correlated function
values, and $l$ defines the time scale of that correlation.

The rational quadratic function is used here instead as it better
models things that may occur on many time scales\cite{Lasko2013}:

$$k(t,t') = \sigma_n^2\bigg[1+\frac{(t-t')^2}{2\alpha\tau^2}\bigg]^{-\alpha}$$

Gaussian process regression was implemented from algorithm 2.1 of
Rasmussen \& Williams\cite{Rasmussen2004}, copied below:

\begin{align}
  & L=\textrm{cholesky}(K+\sigma_n^2I)\\
  & \pmb{\alpha}=L^\top\backslash(L\backslash\mathbf{y}) \\
  & \bar{f_*}=\mathbf{k_*^\top}\pmb{\alpha} \\
  & \bar{v}=L\backslash\mathbf{k_*} \\
  & \textrm{V}\big[f_*\big]=k(\mathbf{x_*},\mathbf{x_*})-\mathbf{v}^\top\mathbf{v} \\
  & \log p(\mathbf{y}|X)= -\frac{1}{2}\mathbf{y}^\top\pmb{\alpha}-\sum_i\log L_{ii}-\frac{n}{2}\log2\pi
\end{align}

$X$ is the training inputs (a $n\times 1$ matrix for a time-series of
$m$ samples), $y$ is a $n\times 1$ matrix with the corresponding
values to $X$, and $\mathbf{x_*}$ is a test input (a scalar), for
which $\bar{f_*}$ are $\textrm{V}\big[f_*\big]$ the predictions
(respectively, predictive mean and variance).

$K$ is a $n\times n$ matrix for which $K_{ij}=k(X_i,X_j)$,
$\mathbf{k_*}$ is a $m\times 1$ matrix for which
$(\mathbf{k_*})_i=k(X_i, \mathbf{x_*})$, and $A\backslash B$ (both $A$
and $B$ matrices) is the matrix $x$ such that $Ax=B$.

Matrices $L$ and $\pmb{\alpha}$ in effect represent the Gaussian
process itself (alongside the covariance function and its
hyperparameters), and can be reused for any test inputs
$\mathbf{x_*}$.  This implementation also exploits the fact that the
algorithm trivially generalizes to multiple $\mathbf{x_*}$ in matrix
form and produces multiple $\bar{f_*}$ and $\textrm{V}\big[f_*\big]$.

Line 6 provides the log marginal likelihood of the target values
$\mathbf{y}$ given the inputs $X$, and this was the basis for
optimizing the hyperparameters of the covariance function. In
specific, in order to optimize the hyperparameters $\sigma_n$, $\tau$,
and $\alpha$, every individual time-series was transformed with the
time-warping described in the prior section, standardized to mean of 0
and standard deviation of 1 (note that the mean and standard deviation
must be saved in order to undo this transformation when
interpolating), and then $\sum\log p(\mathbf{y}|X)$ over the entire
training set was maximized using a grid search.

Hyperparameter optimization is likely the most time-consuming step of
processing, and Gaussian process regression is very sensitive to their
values.  However, this step also is a highly parallelizable one, and
so it was amenable to the distributed nature of Apache Spark (and
likely to more efficient methods such as gradient descent).

The values (i.e. $y$) in each individual time-series were standardized
to a mean of 0 and standard deviation of 1, and this standardization
was then reversed on the interpolated values.

\subsubsection{Interpolation}

The remainder of algorithm 2.1 is not reproduced here, but the code
directly implemented this method using the rational quadratic function
(and hyperparameters given above) as the covariance function.  This
inferred for each individual time-series a continuous function
producing predictive mean and variance for any input $t$ (for which
they use the notation $\mathbf{x^*}$ for ``test input'').

As in \cite{Lasko2013}, all of these inferred functions were then
evaluated at a regular sampling of time values (i.e. via the test
input $\mathbf{x^*}$) with padding added before and after each time
series.  The sampling frequency and the amount of padding depends on
the dataset, and so can be specified when running the tool.

In effect, this mapped each individual time-series first to a
continuous function, and then to a new ``interpolated'' time-series
containing predicted mean and variance at the sampled times described
above.  The interpolated time-series first had the reverse
transformation applied from their standardization (i.e. the stored
mean was added back in), and this was then written to CSV files and
used as input to later steps.

\subsection{Feature Learning with Autoencoder}

A stacked sparse 2-layer autoencoder was then used to perform feature
learning.  The implementation used here was a combination of what was
described in \cite{Lasko2013} (which closely follows the UFLDL
Tutorial from Stanford\cite{Ng}), and Fran\c cois Chollet's
guide\cite{Chollet} on using the Python library
\href{https://keras.io/}{Keras} to implement autoencoders.
Specifically, Keras was used with
\href{http://deeplearning.net/software/theano/}{Theano} (GPU-enabled)
as the backend; all data was loaded from the prior step with the
\href{http://pandas.pydata.org/}{pandas} library.

These autoencoders have a fixed-size input and output, and in order to
accomodate this, fixed-size contiguous patches were sampled from the
interpolated time-series.  As in \cite{Lasko2013}, the patch size was
set to the total number of padded samples, and patches were sampled
uniformly randomly from all contiguous patches.  Note that since
Gaussian process regression produces both mean and variance
predictions, the input and output size of the network are twice the
patch size.

As in \cite{Lasko2013}, the encoder and decoder layers used sigmoid
encoders and linear decoders, and all hidden layers had 100 units.
Both layers contained a sparsity constraint (in the form of L1
activity regularization) and L2 weight regularization, and performance
appeared very sensitive to their weights.

\begin{figure}
  \includegraphics[width=0.7\linewidth]{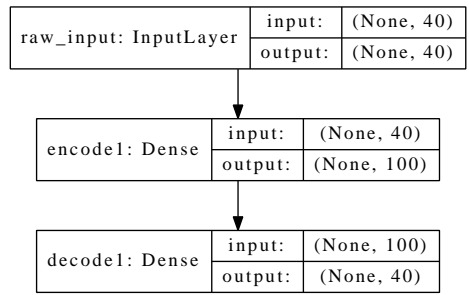}
  \caption{First stage of Keras autoencoder}
  \label{fig:keras_autoencoder1}
\end{figure}

\begin{figure}
  \includegraphics[width=0.7\linewidth]{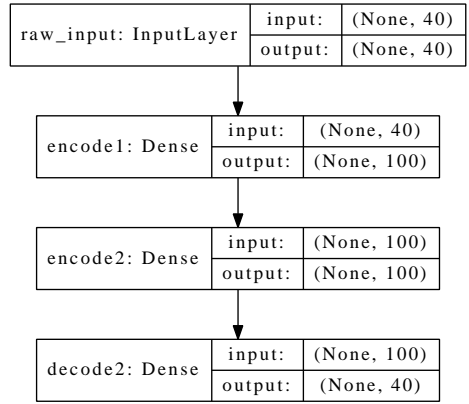}
  \caption{Second stage of Keras autoencoder}
  \label{fig:keras_autoencoder2}
\end{figure}

This network was built up in stages in order to greedy layerwise
train\cite{Ng}.  The first stage was the neural network in figure
\ref{fig:keras_autoencoder1}; this was trained with the raw input data
(at both the input and the output), thus learning ``primary'' hidden
features at the layer \texttt{encode1}.  The first decoder layer
\texttt{decode1} was then discarded and the model extended with
another encoder and decoder, as in figure
\ref{fig:keras_autoencoder2}.

This model then was similarly trained on raw input data, but while
keeping the weights in layer \texttt{encode1} constant.  That is, only
layers \texttt{encode1} and \texttt{decode2} were trained, and in
effect they were trained on ``primary'' hidden features
(i.e. \texttt{encode1}'s activations on raw input).  Following this
was ``fine-tuning''\cite{Ng} which optimized all layers, i.e. the
stacked autoencoder as a whole, again using the raw input data.

The final model then discarded layer \texttt{decode2}, and used the
activations of layer \texttt{encode2} as the learned sparse features.
(Elsewhere in the paper, ``second-layer learned features'' refers to
these activations on a given patch of time-series input.  ``First
layer learned features'' refers to the activations of
\texttt{encode1}.)


\subsection{Visualization \& Classification}

The final processing of the tool consists of using the learned sparse
features of the prior step (from both the first and second layers) as
input into two separate steps: a visualization by way of t-SNE
(t-distributed Stochastic Neighbor Embedding), and training a logistic
regression classifier.

Both of these steps used the Python library
\href{http://scikit-learn.org/stable/index.html}{scikit-learn}, and
respectively
\href{http://scikit-learn.org/stable/modules/generated/sklearn.manifold.TSNE.html}{sklearn.manifold.TSNE}
and
\href{http://scikit-learn.org/stable/modules/generated/sklearn.linear\_model.LogisticRegression.html}{sklearn.linear\_model.LogisticRegression}.

\section{Experimental Evaluation}
\label{expeval}

As a test of the tool described in this paper, experiments were run on
a selection of the data.  Particularly, the LOINC code
\texttt{1742-6}, corresponding to MIMIC-III \texttt{ITEMID} of 50861,
Alanine Aminotransferase (ALT), was used, and the ICD-9 categories 428
(heart failure) and 571 (chronic liver diseases), corresponding to
ALT's use as a biomarker for liver health and to suggest congestive
heart failure.  All time-series for this were in international
units/liter (IU/L).

This were selected to a total of 3,553 unique admissions (1,782 for
ICD-9 category 428, and 1,771 for 571), 3,320 patients (1,397 females,
1,923 males), and 34,047 time-stamped samples.  70\% of these
admissions were randomly selected for the training set, and the
remaining 30\% for the testing set (2,473 and 1,080 respectively).
The interpolated series from Gaussian process regression were padded
by 10 samples at the beginning and end and sampled at 0.25 days (thus
2.5 days of padding), producing a total of 198,830 samples.

These interpolated time-series were randomly resampled to 7,419
patches of 20 samples long (thus, the neural network used inputs and
outputs of 40 nodes).  20\% of these were set aside for
cross-validation.

\begin{figure}
  \includegraphics[width=\linewidth]{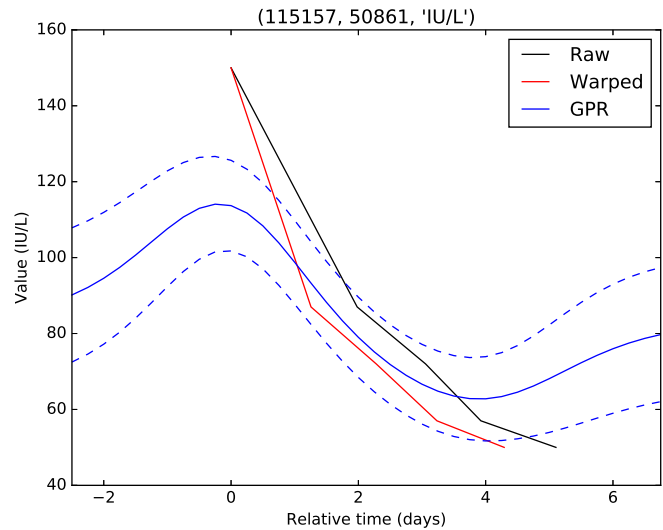}
  \caption{Time-series: Original, warped, and GPR interpolated}
  \label{fig:ts_gpr1}
\end{figure}

\begin{figure}
  \includegraphics[width=\linewidth]{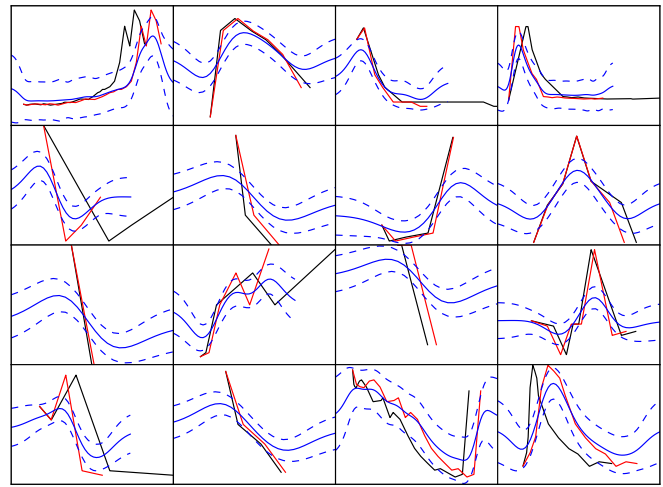}
  \caption{Example time-series}
  \label{fig:ts_gpr2}
\end{figure}

Figure \ref{fig:ts_gpr1} is an example of a time-series from this
data.  The solid black line is the ``original'' time-series, the red
line is the warped version, and the blue line is the version after
interpolation with Gaussian process regression (with one standard
deviation plotted on the surrounding dotted line).  Figure
\ref{fig:ts_gpr2} is several other randomly-chosen time-series from
the data as examples.

When training the autoencoder on this data, manual tuning led to an L1
activity regularization (i.e. sparsity constraint) of $10^{-4}$
and L2 weight regularization of $10^{-3}$.

\begin{figure}
  \includegraphics[width=\linewidth]{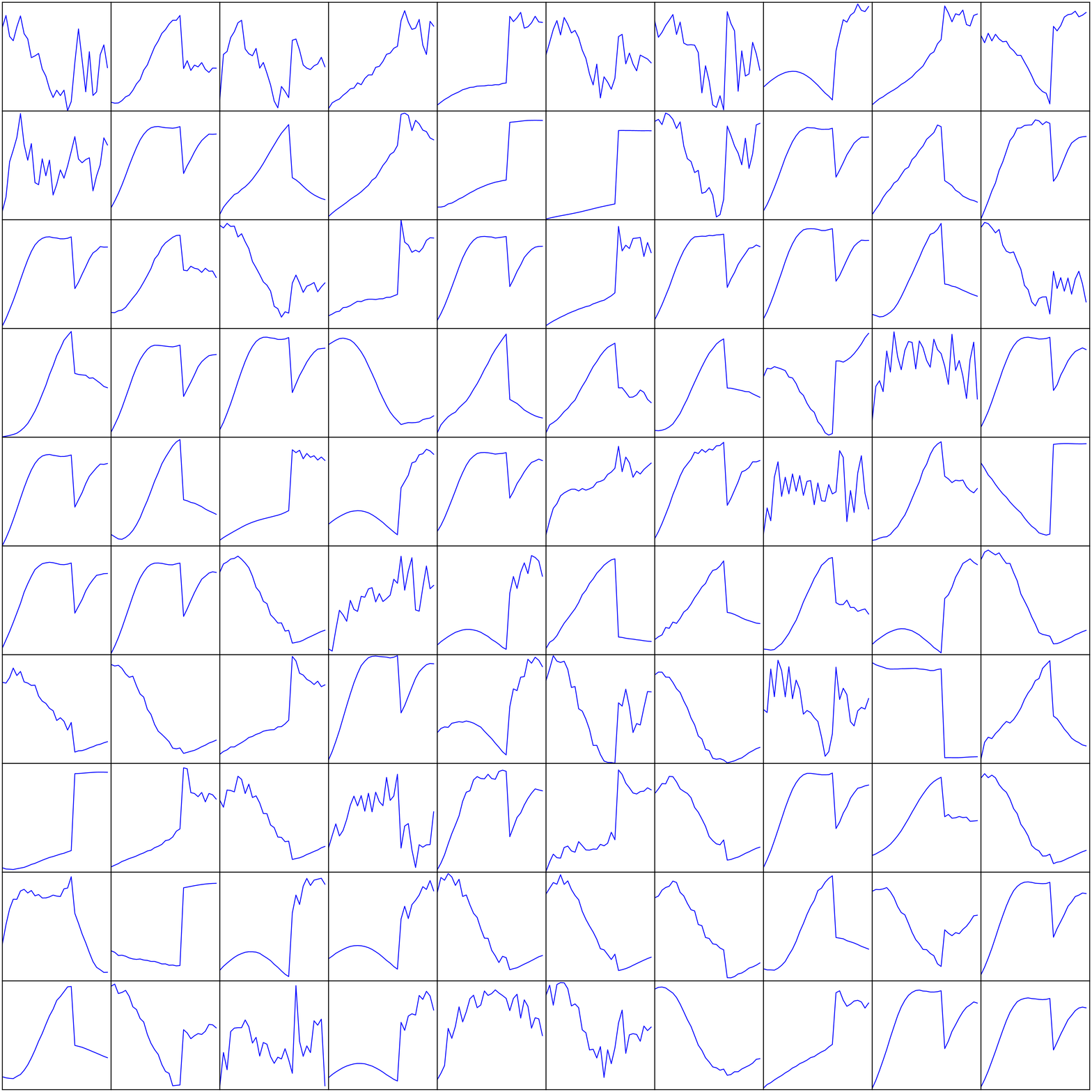}
  \caption{Autoencoder first-layer weights, shown as 100 time-series}
  \label{fig:keras_hidden1}
\end{figure}

An interesting detail which figure 2 of \cite{Lasko2013} shows is that
the effects of the first layer can be visualized directly in the form
of its weights (not its activations; this is the result of its
training, not of any input).  As they form a 40x100 array, each
40-element vector corresponds to a sort of time-series signature which
that unit is detecting.  Figure \ref{fig:keras_hidden1} shows the
corresponding plot of first-layer weights (i.e. \texttt{encode1})
after it is trained on the subset described here.

This shows similar structure as in \cite{Lasko2013} (including
considerable redundancy), but with different sorts of signatures.
Particularly, it seems to single out edges and certain kinds of ramps.

The t-SNE results, shown below in figures \ref{fig:tsne1} and
\ref{fig:tsne2}, are inconclusive here.  It appears to have extracted
some structure (which in the second layer is better-refined), but this
structure does not seem to relate clearly with the labels.

\begin{figure}
  \includegraphics[width=\linewidth]{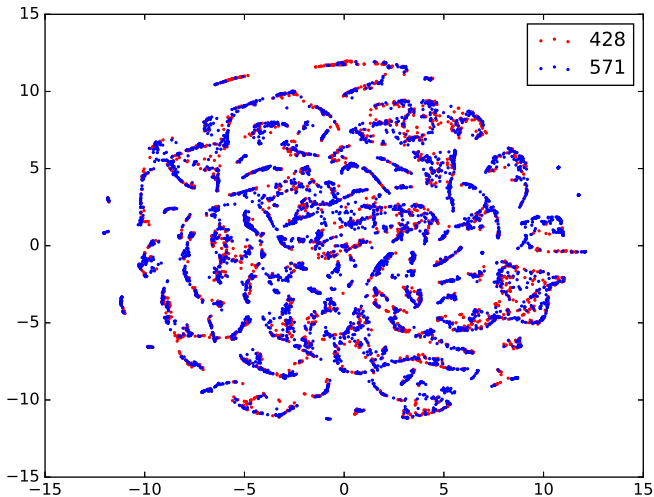}
  \caption{t-SNE on first-layer learned features}
  \label{fig:tsne1}
\end{figure}

\begin{figure}
  \includegraphics[width=\linewidth]{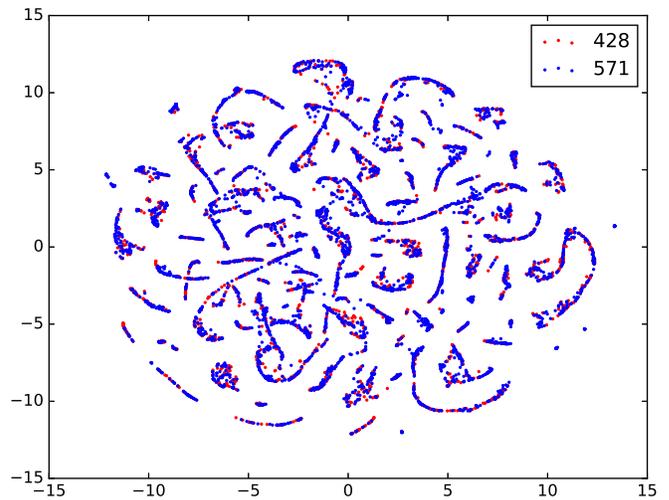}
  \caption{t-SNE on second-layer learned features}
  \label{fig:tsne2}
\end{figure}

The classifier here is not producing useful results; particularly, it
is producing an AUC of 0.5 on both the 1st and 2nd layer features.
The reason for the poor performance is not known, but due to this, it
was not compared against any ``baseline'' classifier or expert-feature
classifier as in \cite{Lasko2013}.  Overall, further work is needed
here.

\section{Conclusions \& Further Work}

The tool is released as open source built on openly available
libraries and (mostly) open data sources.  It was sufficient to
produce all diagrams, plots, and analysis in this paper.  However, it
still needs further experimentation to produce meaningful results, and
the intention is that it can be a starting point for this.

The examples were restricted by the tool's use of ICD-9 categories,
which may have been too broad to produce meaningful clustering.
Generalizing this would be useful, as would other diagnostics to give
clues into the feature learning process (such as plots of second-layer
learned features).

The original goal of running as much as possible within Apache Spark
on ``standard'' infrastructure such as
\href{https://aws.amazon.com/emr/}{Amazon EMR} or
\href{https://databricks.com/}{Databricks} was not fully met.  Further
integration with Apache Spark still is possible; the autoencoders
perhaps could be implemented in
\href{https://deeplearning4j.org/}{DL4J} (a native Java library
supporting Apache Spark) or Spark's built-in \texttt{pyspark} support
may allow the Keras and scikit-learn code to run directly on that
infrastructure via \texttt{spark-submit}.  The R language also has
many relevant libraries, and
\href{https://spark.apache.org/docs/latest/sparkr.html}{SparkR} may at
some point permit their more seamless use.

Several optimizations also may help.  hyperparameter optimization is
currently done with a grid search, but would more sensibly done with a
more intelligent optimization algorithm (such as SGD).  The time
warping function has parameters that could be tuned, or more extensive
changes\cite{Lasko2015} could be made to try to make the time-series
more stationary.  Other covariance functions may be more appropriate
as well.

Some other areas should perhaps be explored further too.  One
incremental change is in the use of multiple-task Gaussian processes
(MTGPs); the work done here handles only individual time-series, while
MIMIC-III is rich in parallel time-series that correlate with each
other.  Ghassemi \emph{et al.}\cite{Ghassemi2015} explored the use of
MTGPs to find a latent representation of multiple correlated
time-series, but did not use this representation for subsequent
feature learning.  Another incremental change is in the use of
Variational Autoencoders (VAEs) to learn a feature space that is
sufficiently low-dimensional that techniques such as t-SNE are not
required for effective visualization.

A more extensive change could involve using recurrent neural networks
(RNNs).  Deep networks such as RNNs such as Long Short-Term Memories
(LSTMs) have shown promise in their ability to more directly handle
sequences\cite{Sutskever2014} and clinical time-series data, including
handling missing data\cite{Lipton2015, Lipton2015a, Lipton2016}.
However, they are primarily used for supervised learning, but could
potentially be treated similarly as autoencoders (as in
\cite{Klapper-Rybicka2001}), that is, trained with the same input and
output data in order to learn a reduced representation of the input.
This approach would avoid some of the need to perform Gaussian Process
Regression, however, it still may not cope well with time-series data
that is very irregular.

\section*{Acknowledgment}
Thank you to Dr. Jimeng Sun, Sungtae An, and the other TAs for their
time and advice in this project.

\bibliography{report,report_websites}
\bibliographystyle{abbrv}

\end{document}